\documentclass[letterpaper,10pt,conference]{ieeeconf}
\IEEEoverridecommandlockouts
\usepackage{afterpage}
\usepackage{amsmath}
\usepackage{amssymb}
\usepackage{booktabs}
\usepackage{graphicx}
\usepackage{multirow}
\usepackage{array}
\usepackage{tabularx}
\usepackage{pifont}
\usepackage{xcolor}
\usepackage{xspace}
\usepackage{url}
\usepackage{siunitx}
\usepackage[colorlinks=true,
            linkcolor=blue,
            citecolor=blue,
            urlcolor=blue]{hyperref}

\newcommand{\method}{\textit{Radar2Plan}\xspace}
\newcommand{\cmark}{\ding{51}}
\newcommand{\xmark}{\ding{55}}

\title{\LARGE \bf Radar2Plan: Benchmarking 4D Radar for End-to-End Open-Loop Ego-Trajectory Planning}

\author{
Ling Yao$^{1,2}$,
Yichun Xiao$^{1}$,
Jin Jin$^{3}$,
Yihan Zhang$^{2}$,
Fangqiang Ding$^{1,*}$
\\[0.6em]
{\small
$^{1}$HKUST(GZ)
\quad
$^{2}$Shanghai Jiao Tong University
\quad
$^{3}$University of Oxford
}
\thanks{$^{*}$ Corresponding author.}
}

\begin{document}
\bstctlcite{BSTcontrol}
\maketitle
\thispagestyle{empty}
\pagestyle{empty}

\begin{abstract}
Adverse weather and poor illumination remain major challenges for robust ego-trajectory planning in mobile autonomy.
4D radar offers reliable sensing under adverse conditions and direct radial-velocity measurements.
However, sensing robustness does not necessarily translate into robust downstream planning, while existing 4D radar benchmarks focus primarily on perception rather than trajectory planning.
We present \method, a modular benchmark for open-loop ego-trajectory planning using real-world 4D radar data.
\method connects sensor encoders, scene representations, and planning heads through common interfaces, enabling controlled comparisons between different sensor and planner configurations.
Using DSERT-RoLL and MAN TruckScenes, we evaluated seven combinations of camera, 4D radar, and LiDAR in four representative planning baselines and 12 weather and illumination conditions under a unified protocol.
To our knowledge, \method is the first benchmark dedicated to evaluating real-world 4D radars for autonomous driving planning.
Experiments show that 4D radar alone supports competitive ego-trajectory planning and robust performance under adverse conditions.
Sensor-configuration comparisons further demonstrate its complementary value to other modalities, while revealing dependencies on sensor combination and planning architecture.
The modular design also supports additional datasets, sensing modalities, and planners, providing a flexible foundation for future research on radar-based autonomous driving planning.

\end{abstract}

\begin{table*}[!t]
\centering
\caption{
Comparison of representative planning benchmarks and 4D radar datasets.}
\label{tab:benchmark_compare}
\setlength{\tabcolsep}{4.5pt}
\resizebox{\textwidth}{!}{
\begin{tabular}{lccccccc}
\toprule
Benchmark & Real sensor logs & Real 4D radar & Ego planning & Long range & Condition eval. & Modality controls & Closed loop \\
\midrule
nuScenes~\cite{caesar2020nuscenes}
    & \cmark & \xmark & \xmark & \xmark & \xmark & \xmark & \xmark \\
nuPlan~\cite{caesar2021nuplan}
    & \cmark & \xmark & \cmark & \xmark & \xmark & \xmark & \cmark \\
Waymax~\cite{gulino2023waymax}
    & \xmark & \xmark & \cmark & \xmark & \xmark & \xmark & \cmark \\
Bench2Drive~\cite{jia2024bench2drive}
    & \xmark & \xmark & \cmark & \xmark & \cmark & \xmark & \cmark \\
\midrule
K-Radar~\cite{paek2022kradar}
    & \cmark & \cmark & \xmark & \xmark & \cmark & \xmark & \xmark \\
View-of-Delft~\cite{palffy2022vod}
    & \cmark & \cmark & \xmark & \xmark & \xmark & \xmark & \xmark \\
TJ4DRadSet~\cite{zheng2022tj4dradset}
    & \cmark & \cmark & \xmark & \xmark & \xmark & \xmark & \xmark \\
DSERT-RoLL~\cite{cho2026dsert}
    & \cmark & \cmark & \xmark & \xmark & \cmark & \xmark & \xmark \\
MAN TruckScenes~\cite{fent2024truckscenes}
    & \cmark & \cmark & \xmark & \cmark & \xmark & \xmark & \xmark \\
TruckDrive~\cite{ghilotti2026truckdrive}
    & \cmark & \cmark & \cmark & \cmark & \xmark & \xmark & \xmark \\
\midrule
\textbf{\method}
    & \cmark & \cmark & \cmark & \cmark & \cmark & \cmark & \xmark \\
\bottomrule
\end{tabular}
}
\end{table*}

\section{Introduction}
\label{sec:intro}

\begin{figure}[t]
\centering
\includegraphics[width=\columnwidth]{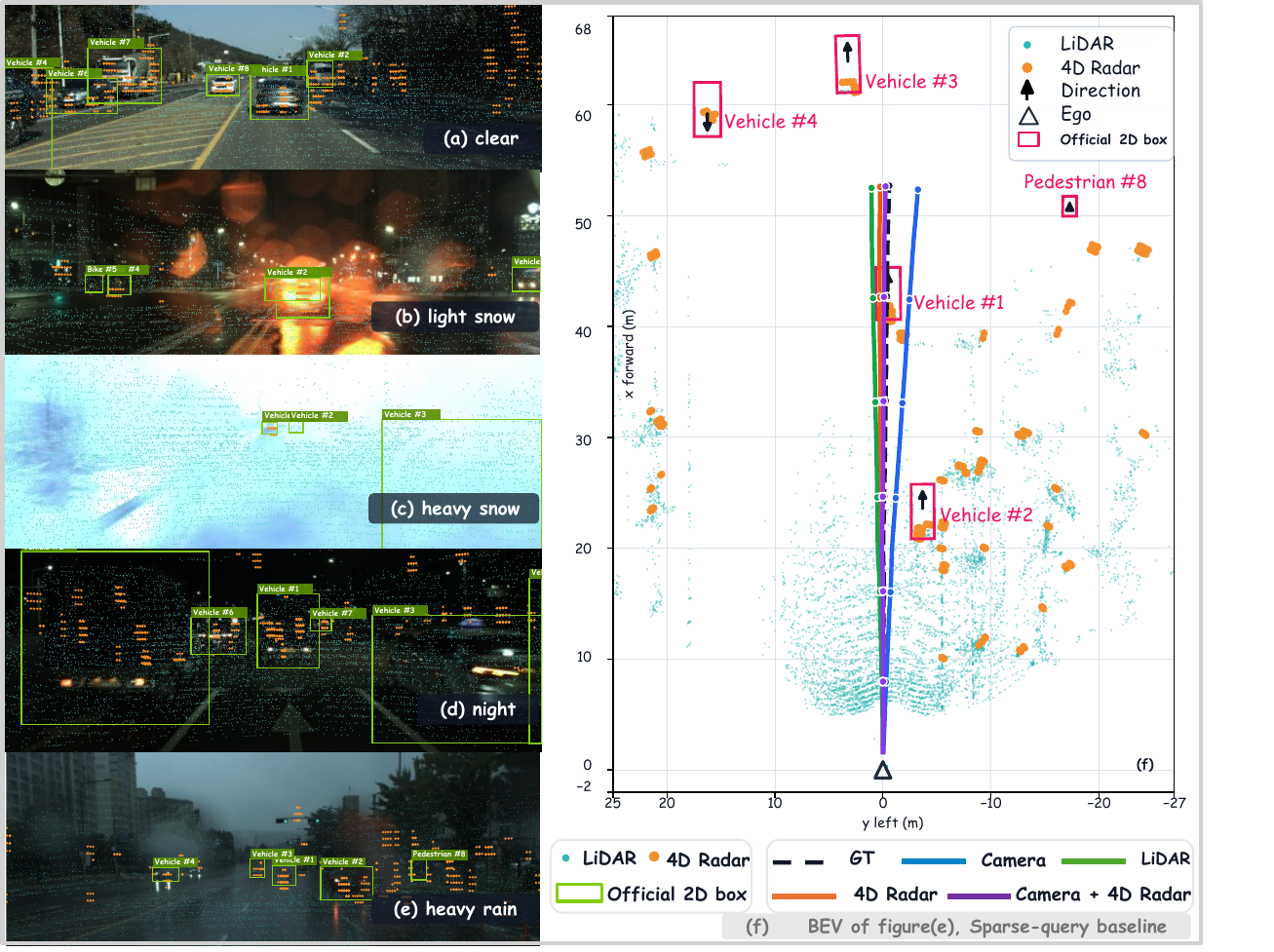}
\caption{
Qualitative illustration of how 4D radar sensing robustness benefits downstream planning.
(a)--(e) Projected 4D radar returns remain reliable or hardly affected under clear, snow, night, and heavy rain conditions.
(f) Ego-frame bird's-eye-view (BEV) of the heavy rain scene in (e).
A SparseDrive-style sparse-query planner~\cite{sun2025sparsedrive} predicts Top-1 trajectories in camera-only, radar-only, LiDAR-only, and camera-radar sensor configurations.
The radar-only and camera-radar predictions remain close to the logged trajectory despite degraded visual observations.
}
\label{fig:teaser}
\end{figure}

Mobile autonomous systems, including autonomous cars and trucks, UAVs, and construction machines, operate in safety-critical environments and must plan reliable motion from sensor observations.
Cameras and LiDAR are widely used for environmental sensing, but cameras can be severely affected by poor illumination and adverse weather, while LiDAR measurements can also degrade in conditions such as rain, fog, and snow~\cite{han20244dradarsurvey,paek2022kradar}, making robust motion planning challenging.
In comparison, the 4D radar is less sensitive to these conditions and directly measures range, azimuth, elevation, and radial velocity~\cite{paek2022kradar,zheng2022tj4dradset}, making it a promising sensing modality for robust planning of autonomous driving.
Fig.~\ref{fig:teaser} illustrates the
robustness of 4D radar sensing under adverse conditions against camera and LiDARs and its benefits to downstream trajectory planning.

Despite these advantages, existing 4D radar studies have focused primarily on perception tasks such as object detection, tracking, and occupancy prediction~\cite{paek2022kradar,palffy2022vod,zheng2022tj4dradset}, leaving its value for downstream planning insufficiently evaluated.
This gap has become increasingly relevant as autonomous driving systems have shifted from modular perception-prediction-planning pipelines toward end-to-end models~\cite{hu2023uniad,jiang2023vad} that directly predict ego trajectories from sensor observations.
In such systems, the quality and robustness of the sensing modality can directly influence the resulting planning performance.
However, whether the sensing robustness of 4D radar translates into more robust ego-trajectory planning has not been systematically evaluated.

This motivates our central research question: \emph{Does the sensing robustness of 4D radar benefit downstream planning?}
We investigate this question from three complementary perspectives:
1) can a planner generate competitive ego trajectories using 4D radar as its only environmental sensing modality?
2) under adverse conditions, does radar-based planning achieve lower planning error than camera or LiDAR-based planning?
and 3) does adding 4D radar improve planning when combined with camera, LiDAR, or camera-LiDAR sensor inputs?
These questions examine the radar-only planning performance, the robustness of radar-based planning under adverse conditions, and its complementarity with other sensing modalities.

We therefore present \method, to our knowledge, the first benchmark for open-loop ego-trajectory planning using real-world 4D radar data.
Supporting heterogeneous datasets, sensing modalities, and planning methods requires handling differences in data organization, sensor representations, and planner-specific interfaces.
If treated independently for each experimental configuration, these differences can introduce implementation-specific variations and hinder controlled comparisons.
To address this, \method adopts a modular design that separates dataset adaptation, sensor input encoding, planning-specific scene representation, and the planning method through standardized interfaces.
This design allows individual components to be exchanged independently while maintaining a consistent evaluation protocol across sensor configurations and planning methods.
\method evaluates seven combinations of camera, 4D radar, and LiDAR in four representative planning baselines: GRU-based waypoint regression adapted from TransFuser~\cite{prakash2021transfuser}, sparse-query planning adapted from SparseDrive~\cite{sun2025sparsedrive}, trajectory-vocabulary scoring adapted from VADv2~\cite{jiang2026vadv2} and diffusion-based trajectory refinement adapted from DiffusionDrive~\cite{liao2025diffusiondrive}.

Across both DSERT-RoLL and MAN TruckScenes, planners using 4D radar as the sole sensing modality consistently outperform their camera and LiDAR-only counterparts.
Adding 4D radar also generally improves multi-modal sensing-based planning, with particularly pronounced benefits under adverse weather and illumination conditions.
These results demonstrate that the sensing robustness of 4D radar can translate into improved downstream ego-trajectory planning, and further highlight the importance of a dedicated benchmark for systematically evaluating this transfer across different sensor configurations, environmental conditions, and planning architectures.

Our main contributions are summarized as follows:
\begin{itemize}

\item We formulate and systematically investigate whether the sensing robustness of real-world 4D radar translates into improved open-loop ego-trajectory planning.
To enable this study, we introduce \method, which converts DSERT-RoLL and MAN TruckScenes into a unified planning benchmark with consistent task definitions, data splits and evaluation metrics.

\item We develop a modular design of the benchmark that enables controlled comparisons across datasets, sensing modalities, planner-specific scene representations, and planning methods, while allowing individual components to be changed or extended independently.

\item
Through extensive evaluations across multiple sensor configurations and planning baselines, we show that 4D radar provides robust support for ego-trajectory planning, particularly under adverse sensing conditions.
The configuration-dependent results further highlight the importance of a dedicated benchmark for systematically assessing the planning value of 4D radar across different sensor combinations and planning architectures.

\end{itemize}

\begin{figure*}[t]
\centering
\includegraphics[width=\textwidth]{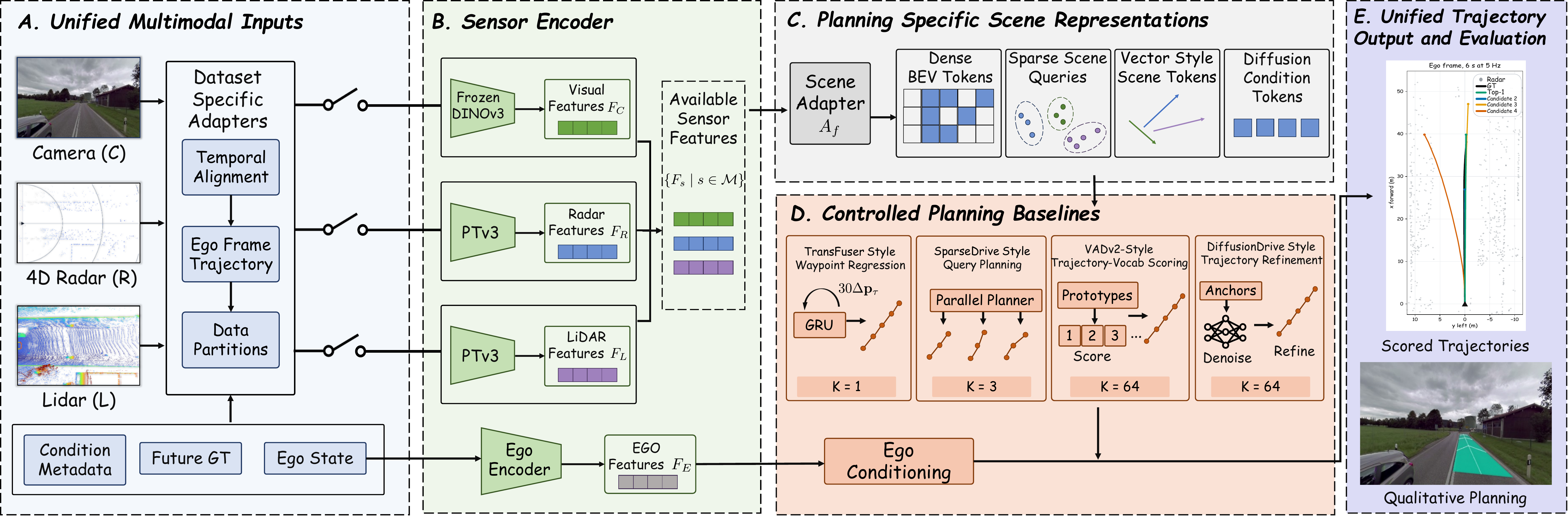}
\caption{Overview of \method.
(A) Dataset-specific adapters construct planning samples for seven configurations of camera ($C$), 4D radar ($R$), and LiDAR ($L$).
(B)-(D) Modality-specific encoders and planning-specific scene adapters connect these inputs to four representative planning baselines.
(E) A common interface returns scored trajectory candidates for evaluation.
}
\label{fig:benchmark_overview}
\end{figure*}

\section{Related Work}

\subsection{4D Radar Perception and Datasets}

Compared to conventional automotive radar, the 4D radar provides elevation-aware measurements and improved angular resolution while retaining robustness to poor illumination and adverse weather~\cite{han20244dradarsurvey,paek2022kradar}.
These properties have motivated a growing number of 4D radar datasets and perception benchmarks.
K-Radar~\cite{paek2022kradar}, View-of-Delft~\cite{palffy2022vod}, TJ4DRadSet~\cite{zheng2022tj4dradset}, and DSERT-RoLL~\cite{cho2026dsert}, for example, primarily evaluate 4D radar through tasks such as 3D object detection and multimodal perception.
Radar-camera methods including CenterFusion~\cite{nabati2021centerfusion}, CRN~\cite{kim2023crn}, RCBEVDet~\cite{lin2024rcbevdet}, RaCFormer~\cite{chu2025racformer} and ZFusion~\cite{yang2025zfusion} further demonstrate that radar features can complement visual observations for better 3D scene understanding.
Collectively, these studies establish the sensing and perception value of 4D radar, but do not determine whether these advantages translate into improved downstream ego-trajectory planning.

Evaluating this question requires real-world datasets that provide synchronized 4D radar together with other commonly used sensing modalities and accurate ego-motion information.
Among existing datasets, DSERT-RoLL~\cite{cho2026dsert} and MAN TruckScenes~\cite{fent2024truckscenes} satisfy these requirements by providing synchronized and calibrated camera, 4D-radar, LiDAR, and ego-motion measurements.
They also cover complementary driving scenarios: DSERT-RoLL contains urban and suburban driving under diverse weather and illumination conditions, while MAN TruckScenes includes heavy-truck operation on motorways, feeder roads, and terminal roads.
These properties allow future ego trajectories to be constructed from real-world recordings and enable controlled comparisons among camera, radar, and LiDAR-based sensor configurations across different operating conditions.
We therefore use these two datasets to study whether the sensing robustness of 4D radar transfers to downstream planning.

\subsection{End-to-End Ego-Trajectory Planning}

End-to-end autonomous driving methods predict future ego trajectories directly from sensor observations~\cite{hu2022stp3,hu2023uniad,jiang2023vad}.
Existing approaches differ substantially in their scene representations and trajectory-generation mechanisms.
TransFuser, TCP, and ThinkTwice employ direct waypoint or trajectory regression~\cite{prakash2021transfuser,wu2022tcp,jia2023thinktwice}.
UniAD, VAD, and SparseDrive instead couple scene understanding and planning through learned queries, vectorized representations, or sparse scene representations~\cite{hu2023uniad,jiang2023vad,sun2025sparsedrive}.
More recent methods explicitly model multiple possible ego trajectories:
VADv2 scores candidates from a predefined trajectory vocabulary~\cite{jiang2026vadv2}, GenAD learns a generative trajectory distribution~\cite{zheng2024genad}, and DiffusionDrive and GoalFlow generate multi-modal trajectories through truncated diffusion and flow matching, respectively~\cite{liao2025diffusiondrive,xing2025goalflow}.

Despite this diversity in planning architectures, most existing methods rely primarily on cameras, LiDAR, or their combination for environment sensing~\cite{prakash2021transfuser,hu2023uniad,jiang2023vad,sun2025sparsedrive}.
SpaRC-AD~\cite{wolters2025sparcad} is a closely related exception that incorporates conventional radar-camera fusion into a SparseDrive-based framework.
However, it evaluates normal radar within a single planning architecture and does not study 4D radar.
TruckDrive~\cite{ghilotti2026truckdrive} provides real-world 4D radar data and a LiDAR-only UniAD ego-planning baseline.
However, its released planner does not use 4D radar as input.
Consequently, existing works cannot establish whether the planning benefit of 4D radar is consistent across fundamentally different end-to-end trajectory-generation mechanisms.

To distinguish modality-level effects to different planning methods, we evaluate 4D radar across four representative planning paradigms: direct waypoint regression, sparse-query planning, trajectory-vocabulary scoring, and diffusion-based trajectory refinement.
We adopt TransFuser~\cite{prakash2021transfuser}, SparseDrive~\cite{sun2025sparsedrive}, VADv2~\cite{jiang2026vadv2}, and DiffusionDrive~\cite{liao2025diffusiondrive} as representative methods for these four paradigms, respectively.
This selection enables us to examine whether the observed planning value of 4D radar persists across substantially different scene representations and trajectory-generation mechanisms.

As summarized in Table~\ref{tab:benchmark_compare}, existing planning benchmarks and 4D radar datasets address complementary parts of the problem, motivating a unified benchmark to evaluate the planning value of 4D radar across sensor configurations, environmental conditions, and planning architectures.

\section{The \method Benchmark}
\label{sec:benchmark}

As shown in Fig.~\ref{fig:benchmark_overview}, \method uses dataset-specific adapters to convert continuous DSERT-RoLL and MAN TruckScenes recordings into a unified set of planning samples, each defined at a reference timestamp and associated with a future ego trajectory.
For each sensing modality, all planning baselines use the same encoder architecture to ensure consistent feature extraction.
The encoded sensor features are then transformed into the scene representation required by each planning architecture.
Finally, all planners produce scored candidate ego trajectories through a common output interface.

\subsection{Unified Multimodal Inputs}
\label{sec:multimodal_inputs}

\subsubsection{Datasets and Task Statistics}

We use DSERT-RoLL~\cite{cho2026dsert}  and MAN TruckScenes~\cite{fent2024truckscenes} to evaluate \method across different road environments and vehicle platforms.
DSERT-RoLL focuses on urban and suburban driving under six weather conditions.
MAN TruckScenes covers autonomous-trucking scenarios on motorways, feeder and terminal roads.
Together, they allow \method to evaluate 4D radar-based trajectory planning in different road structures, weather conditions, sensing ranges, and ego-motion distributions.

DSERT-RoLL provides approximately 22,000 synchronized multimodal frames.
We use its left and right RGB cameras, 4D radar point clouds, tilted Livox LiDAR, and ego poses.
MAN TruckScenes contains 747 released scenes of approximately 20 seconds each.
We use its left front and right front cameras, all six 4D radars, keyframe point clouds from all six LiDARs, calibration, and ego poses.

\begin{table}[t]
\centering
\caption{Dataset statistics under the \method planning protocol.
}
\label{tab:dataset_statistics}
\setlength{\tabcolsep}{3pt}
\resizebox{\columnwidth}{!}{
\begin{tabular}{lcccc}
\toprule
Dataset & Setting & Samples & train/val/test & condition groups \\
\midrule
DSERT-RoLL
    & Urban/suburban
    & 10,650
    & 7,052 / 1,305 / 2,293
    & 6
    \\
MAN TruckScenes
    & Truck/highway/terminal
    & 16,403
    & 13,113 / 1,643 / 1,647
    & 6
    \\
\bottomrule
\end{tabular}
}
\vspace{1mm}
\vspace{1mm}
\begin{minipage}{\columnwidth}
\footnotesize
\end{minipage}
\end{table}
\subsubsection{Temporal Alignment and Trajectory Construction}

The two datasets differ in file structure, sensor naming, sampling frequency, calibration format, and ego-pose representation.
We therefore implement a dataset-specific adapter for each dataset.
For each planning sample, the anchor time is the timestamp of the current sensor observations.
Each adapter identifies valid anchor times, aligns the corresponding sensor observations with timestamped ego poses, and constructs a six-second future ego trajectory in the anchor ego frame.

For DSERT-RoLL, synchronized sequence frames and their ego-pose records define candidate anchor times.
Each target time is matched to the nearest future ego pose with a tolerance of \SI{0.1}{\second}.
A sample is discarded if any of the 30 future poses cannot be matched.
For MAN TruckScenes, the official 2Hz keyframes define the anchors.
Future ego poses are linearly interpolated at \SI{5}{\hertz}, while sensor observations are associated through the corresponding keyframe token.

We construct each trajectory in the ego-centric frame of its anchor.
Let $\mathbf{T}_{w\leftarrow e_t}\in SE(3)$ denote the transformation from the ego-centric frame at anchor time $t$ to the world frame.
A future ego position $\mathbf{p}^{w}_{t+\tau}$ is transformed into this ego-centric frame as
\begin{equation}
\begin{bmatrix}
\tilde{\mathbf{p}}^{e_t}_{t+\tau} \\
1
\end{bmatrix}
=
\mathbf{T}_{w\leftarrow e_t}^{-1}
\begin{bmatrix}
\mathbf{p}^{w}_{t+\tau} \\
1
\end{bmatrix}.
\label{eq:transform}
\end{equation}
We retain the planar $(x,y)$ coordinates and sample 30 future waypoints at intervals of \SI{0.2}{\second} over six seconds.
Panel A of Fig.~\ref{fig:benchmark_overview} summarizes this unified sample construction.
Table~\ref{tab:dataset_statistics} summarizes settings, number of samples, data splits, and number of condition groups for both datasets.

\subsubsection{Data Partitions and Sensor Configurations}

For DSERT-RoLL, we partition the official training sequences into weather-stratified, sequence-disjoint training and validation subsets with an 85\%/15\% ratio, using a random seed of 42.
The official validation set is retained exclusively for the final test evaluation.
For MAN TruckScenes, we use mutually disjoint training, validation, and test sets that contain 80\%, 10\% and 10\% of the scenes, respectively.

Let $C$, $R$, and $L$ denote camera, 4D radar, and LiDAR, respectively.
We evaluate the seven
sensor configurations
$\{C,R,L,C{+}R,C{+}L,R{+}L,C{+}R{+}L\}$.
These configurations support uni-modal and multi-modal comparisons.

\begin{figure*}[t]
    \centering
    \includegraphics[width=\textwidth]{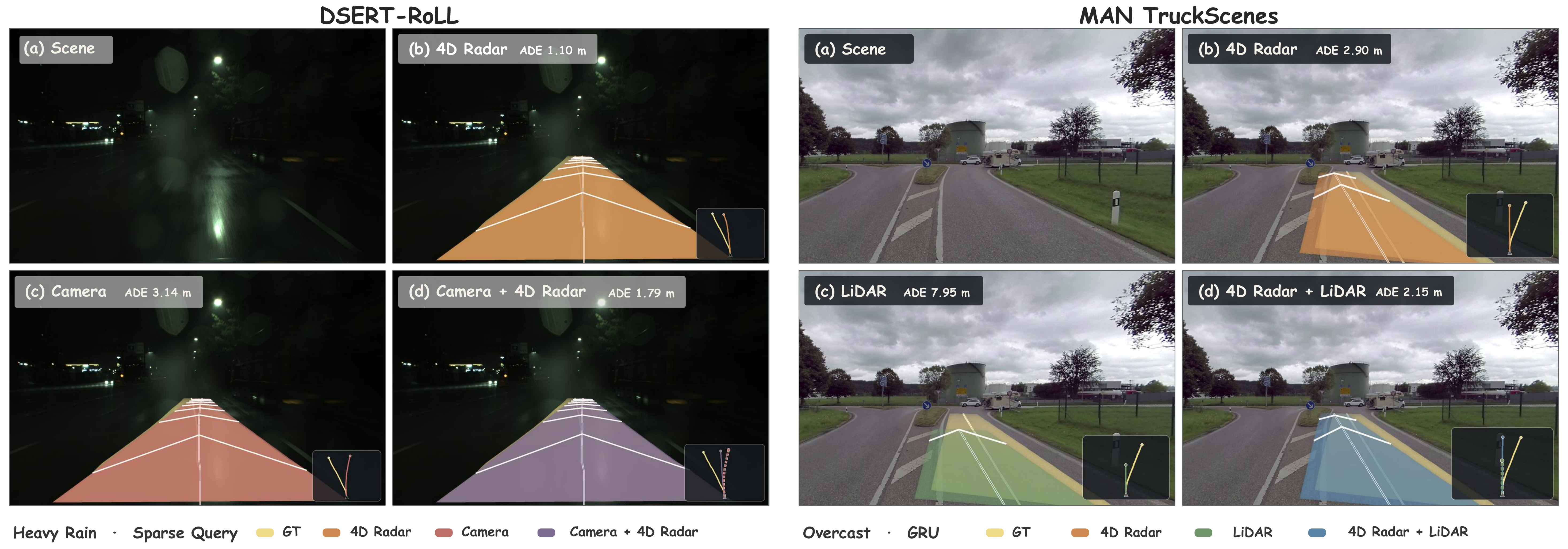}
    \caption{
        Qualitative visualization of 4D-radar-based ego-trajectory planning on DSERT-RoLL under heavy rain and MAN TruckScenes under overcast conditions.
        Insets show the predicted and ground-truth BEV trajectories.
    }
    \label{fig:qualitative_planning}
\end{figure*}

\subsection{Sensor Encoders}
\label{sec:sensor_encoders}

The sensor encoders transform raw observations into feature representations with a common hidden dimension.
We keep the encoder architecture for each sensing modality fixed across all four planning baselines, thereby minimizing confounding effects from differences in encoders of different sensors and enabling a more controlled comparison of the planning architectures.

\subsubsection{Camera Encoder}

We use DINOv3 as the common camera encoder~\cite{simeoni2025dinov3}.
It provides pretrained dense visual features for downstream tasks.
All formal configurations use a frozen DINOv3 ViT-S/16 backbone with the same image preprocessing and an input resolution of $192\times288$.
Only the output projection is trained in each configuration.

\subsubsection{Radar and LiDAR Encoders}

We use separate PTv3-based encoders for 4D radar and LiDAR~\cite{wu2024ptv3}.
Both encoders consist of three PTv3 layers with a patch size of 64 and a grid resolution of \SI{0.5}{\metre}.
The two modalities use separate input projections and independently parameterized encoders to account for their different point-wise attributes.

Before temporal aggregation, each radar point is represented by its ego-frame position $(x,y,z)$, reflectivity, and a Doppler-derived velocity attribute.
For DSERT-RoLL, this velocity attribute corresponds to the provided Doppler measurement, whereas for MAN TruckScenes, it is given by the longitudinal relative velocity expressed in the vehicle's ego-centric frame.
Each LiDAR point is represented by its position $(x,y,z)$, intensity, and relative scan time.

For both radar and LiDAR, the input is formed by aggregating the current point cloud with the preceding frame.
The preceding point cloud is motion-compensated into the current ego frame, and a frame-level temporal offset is appended to each point.
After temporal aggregation, we retain at most 2,048 radar points and 4,096 LiDAR points per sample.

The radar and LiDAR encoders are initialized from scratch for each training run.
Although their architectures are fixed across experiments, their parameters are learned independently for each sensor and planning baseline configuration.

\subsection{Planning-Specific Scene Representations}
\label{sec:scene_representations}

Each planning baseline uses a scene adapter to transform encoded sensor features into the internal representation required by its planning head.
Let $\mathcal{M}\subseteq\{C,R,L\}$ denote a sensor configuration, and let $s\in\mathcal{M}$ denote one sensing modality. Let $f$ denote a planning baseline.
We define
\begin{equation}
\mathbf{F}_s = E_s(\mathbf{X}_s),
\qquad
\mathbf{Z}_{f,\mathcal{M}}
=
A_f\left(\{\mathbf{F}_s \mid s\in\mathcal{M}\}\right),
\label{eq:scene_adapter}
\end{equation}
where $\mathbf{X}_s$ is the input of the sensor modality $s$, $E_s$ is its sensor encoder, and $\mathbf{F}_s$ is the representation of the resulting characteristics.
The scene adapter $A_f$ combines and converts the features of the sensor modalities in $\mathcal{M}$ into $\mathbf{Z}_{f,\mathcal{M}}$.

The four planning baselines use different internal scene representations, reflecting their respective planning architectures.
The TransFuser-style baseline~\cite{prakash2021transfuser} operates on a dense bird's-eye-view (BEV) representation, whereas the SparseDrive-style~\cite{sun2025sparsedrive} and VADv2-style~\cite{jiang2026vadv2} baselines represent the scene using learned sparse queries and latent scene tokens, respectively.
The DiffusionDrive-style baseline~\cite{liao2025diffusiondrive} uses refined BEV features to construct the conditioning tokens required for trajectory generation.

For the TransFuser-style baseline, camera features from multiple views are first aggregated and spatially pooled into a $16\times20$ feature grid.
This grid is then combined with the BEV features obtained from the active radar and LiDAR branches, followed by convolutional fusion and Transformer-based feature encoding.

For the SparseDrive and VADv2-style baselines, the encoded features from all active
sensors are aggregated into a shared sensor memory.
The corresponding sparse queries or latent scene tokens then interact with this memory through cross-attention to obtain planning-relevant scene representations.
For the DiffusionDrive-style baseline, the $16\times20$ BEV features from the active sensor modalities are concatenated and further processed using convolutional spatial refinement to produce the condition tokens used by the diffusion planner.

Each sensor configuration is trained independently and has its
own encoder and scene-adapter parameters.
For a given planning baseline, we keep the scene-adapter architecture and feature dimensions fixed across sensor configurations.
The dense-BEV adapters reserve three $d$-channel $16\times20$ slots for camera, radar and LiDAR features.
An inactive modality is represented by zeros, so the fusion network always
receives $3d$ input channels.
For the query-based methods, tokens from different modalities are concatenated along the sequence dimension.
This changes only the number of tokens, not their feature dimension or the decoder architecture.

\subsection{Controlled Planning Baselines}
\label{sec:planning_baselines}

Using the scene representations described above, \method evaluates four representative trajectory-generation methods under a unified output interface.
To support matched sensor comparisons, we adapt the representative planning components of each baseline to the common benchmark task.
Panel D of Fig.~\ref{fig:benchmark_overview} summarizes their respective planning heads and trajectory outputs.

\subsubsection{TransFuser-Style Waypoint Regression}

The TransFuser-style planner uses the scene representation and ego state to initialize a gated recurrent unit (GRU)~\cite{prakash2021transfuser}.
The GRU sequentially predicts 30 two-dimensional waypoint increments, which are accumulated to obtain the future ego trajectory in the current ego frame.
Since this planner produces a single trajectory, we set $K_{\mathrm{GRU}}=1$.

\subsubsection{SparseDrive-Style Sparse-Query Planning}

The SparseDrive-style planner represents the scene using 64 learned agent queries and 32 learned map queries~\cite{sun2025sparsedrive}.
A planning query interacts with these scene queries, after which the planning head generates three candidate ego trajectories and their mode scores, yielding $K_{\mathrm{SQ}}=3$.

\subsubsection{VADv2-Style Trajectory-Vocabulary Scoring}

The VADv2-style planner constructs 64 fixed trajectory prototypes using only the training set, with the closest prototype serving as the supervision target~\cite{jiang2026vadv2}.
At inference, the planner scores the prototypes and selects the highest-scoring trajectory candidates without residual refinement, yielding $K_{\mathrm{Vocab}}=64$.

\subsubsection{DiffusionDrive-Style Trajectory Refinement}

The DiffusionDrive-style planner initializes 64 trajectory anchors from a vocabulary constructed exclusively from the training set~\cite{liao2025diffusiondrive}.
Conditioned on the scene representation, these anchors are refined through a deterministic two-step truncated-diffusion process with fixed noise.
A separate scene-conditioned scoring head assigns a mode score to each refined trajectory, yielding $K_{\mathrm{Diff}}=64$.

\begin{table*}[t]
\centering
\caption{Overall Top-1 ADE (m) across matched sensor configurations.
Lower values means better results.}
\label{tab:overall_results}

\small
\renewcommand{\arraystretch}{1.18}

\begin{tabular*}{\textwidth}{
@{\extracolsep{\fill}}
l
cccc
cccc
@{}
}
\toprule

& \multicolumn{4}{c}{\textbf{DSERT-RoLL}}
& \multicolumn{4}{c}{\textbf{MAN TruckScenes}} \\
\cmidrule(lr){2-5}
\cmidrule(lr){6-9}

Input
& GRU
& Sparse Query
& Vocabulary
& Diffusion
& GRU
& Sparse Query
& Vocabulary
& Diffusion \\
\midrule

$C$
& 2.71
& 2.95
& 3.20
& 3.06
& 5.54
& 5.78
& 5.75
& 7.03 \\

$R$
& 2.55
& \underline{2.45}
& \underline{2.71}
& \underline{2.51}
& \underline{2.10}
& \textbf{1.96}
& \textbf{2.11}
& \textbf{2.18} \\

$L$
& 2.71
& 2.52
& 2.81
& 2.53
& 6.23
& 6.11
& 11.81
& 6.43 \\

\midrule

$C{+}R$
& 2.57
& \textbf{2.38}
& 3.04
& 2.86
& 2.26
& 2.36
& 2.25
& \underline{2.26} \\

$C{+}L$
& 2.76
& 3.24
& 3.06
& 2.91
& 5.21
& 4.67
& 5.70
& 6.96 \\

$R{+}L$
& \underline{2.49}
& 3.26
& \textbf{2.53}
& \textbf{2.41}
& \textbf{1.76}
& \underline{1.98}
& \underline{2.17}
& 2.29 \\

\midrule

$C{+}R{+}L$
& \textbf{2.48}
& 3.08
& 3.37
& 2.60
& 2.56
& 2.28
& 2.44
& 2.40 \\

\bottomrule
\end{tabular*}

\vspace{1mm}
\begin{minipage}{\textwidth}
\footnotesize
$C$, $R$, and $L$ denote camera, 4D radar, and LiDAR.
Bold and underlined values indicate the best and second-best results
within each planner and dataset.
\end{minipage}
\end{table*}

\subsection{Unified Trajectory Output and Evaluation}
\label{sec:evaluation_protocol}

\subsubsection{Common Output and Top-1 Trajectory Selection}

For planning baseline $f$, the common output interface returns candidate ego trajectories and their associated scores:
\begin{equation}
\hat{\mathbf{Y}}
\in
\mathbb{R}^{B\times K_f\times30\times2},
\qquad
\mathbf{s}
\in
\mathbb{R}^{B\times K_f},
\label{eq:output_contract}
\end{equation}
where $B$ is the batch size and $K_f$ is the number of trajectory candidates produced by baseline $f$.
Each candidate consists of 30 two-dimensional waypoints over six seconds.

For sample $i$, $s_{i,k}$ denotes the score assigned to candidate trajectory $k$.
The Top-1 candidate index is
\begin{equation}
k_i^{\mathrm{top1}}
=
\operatorname*{arg\,max}_{1\leq k\leq K_f}
s_{i,k}.
\label{eq:top1_selection}
\end{equation}
We refer to the selected candidate as the Top-1 trajectory.
The TransFuser-style baseline has $K_f=1$.
Its only trajectory is therefore Top-1 by construction, and its score is fixed to zero.
For the other three baselines, a baseline-specific scoring head predicts one score for each trajectory candidate.
The corresponding scoring objectives are described in Sec.~\ref{sec:experimental_setup}.

\subsubsection{Top-1 Average Displacement Error}

For an evaluation horizon $h\in\{2,4,6\}$\,s, we define $T_h=5h$ as the number of evaluated waypoints.
The Top-1 average displacement error at horizon $h$, denoted as Top1ADE@$h$, is
\begin{align}
\mathrm{Top1ADE@}h
&=
\frac{1}{N}
\sum_{i=1}^{N}
\frac{1}{T_h}
\sum_{t=1}^{T_h}
e_{i,k_i^{\mathrm{top1}},t},
\label{eq:top1_horizon_metrics}
\end{align}
where $N$ is the number of evaluation samples and
$e_{i,k,t}$ is the Euclidean displacement error between waypoint $t$ of candidate $k$ and its ground-truth ego-frame position.

Top1ADE@$h$ averages the displacement error of the Top-1 trajectory over all waypoints up to horizon $h$.
We use Top1ADE@6s as the primary evaluation metric.
Unless otherwise specified, Top1ADE refers to Top1ADE@6s.

\subsubsection{Condition-Specific Evaluation}
\label{sec:condition_evaluation}

We report Top1ADE@6s separately for different weather and illumination conditions.
For DSERT-RoLL, we evaluate six weather groups: Clear, Fog, Light Rain, Heavy Rain, Light Snow, and Heavy Snow.
For MAN TruckScenes, we evaluate six metadata-derived condition groups: Clear, Overcast, Rain, Illumination, Twilight, and Dark.

\section{Experiments}
\label{sec:experiments}

\begin{figure*}[t]
    \centering
    \includegraphics[width=\textwidth]{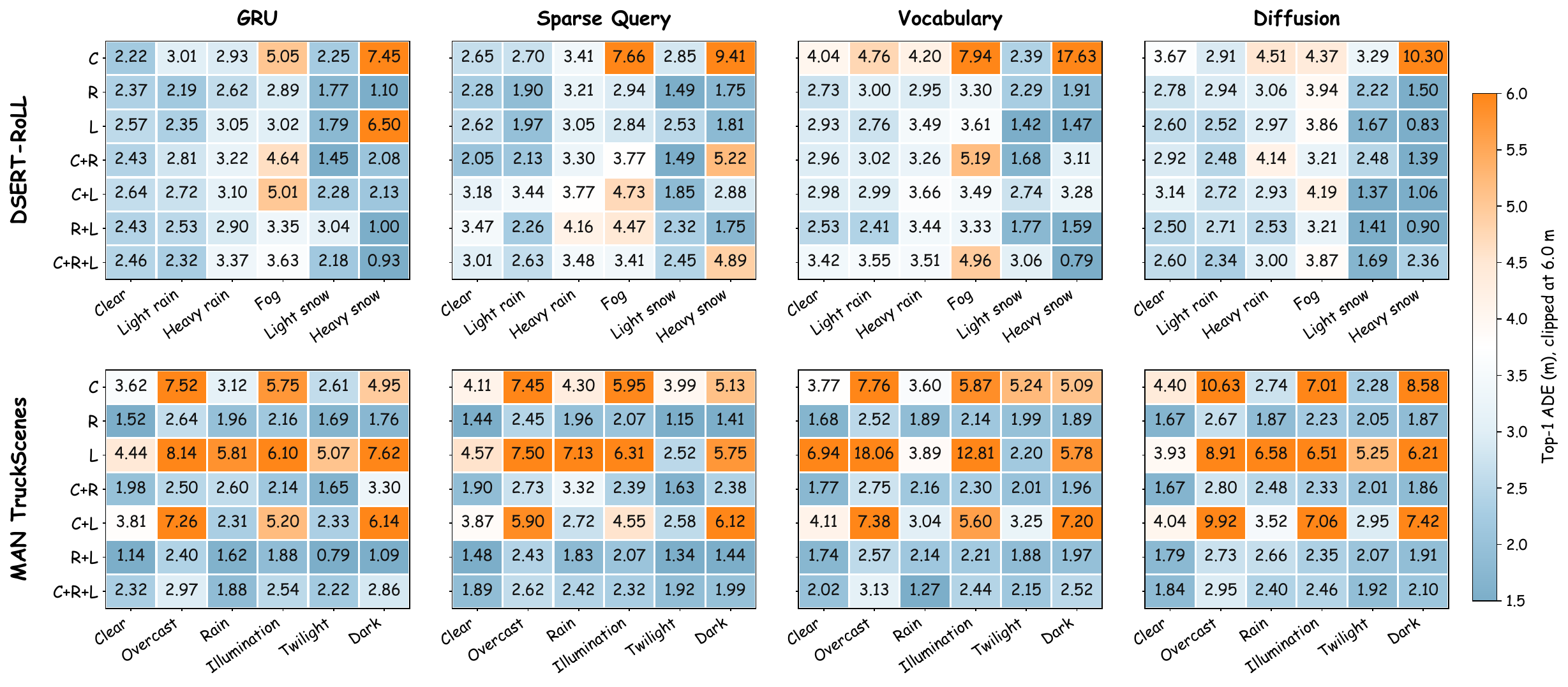}
    \caption{Condition-stratified six-second Top1ADE on
    DSERT-RoLL and MAN TruckScenes. Rows within each dataset denote sensor configurations, and columns denote dataset-specific weather or illumination conditions.}
    \label{fig:condition_specific_radar}
\end{figure*}

\begin{figure}[t]
    \centering
    \includegraphics[width=\columnwidth]{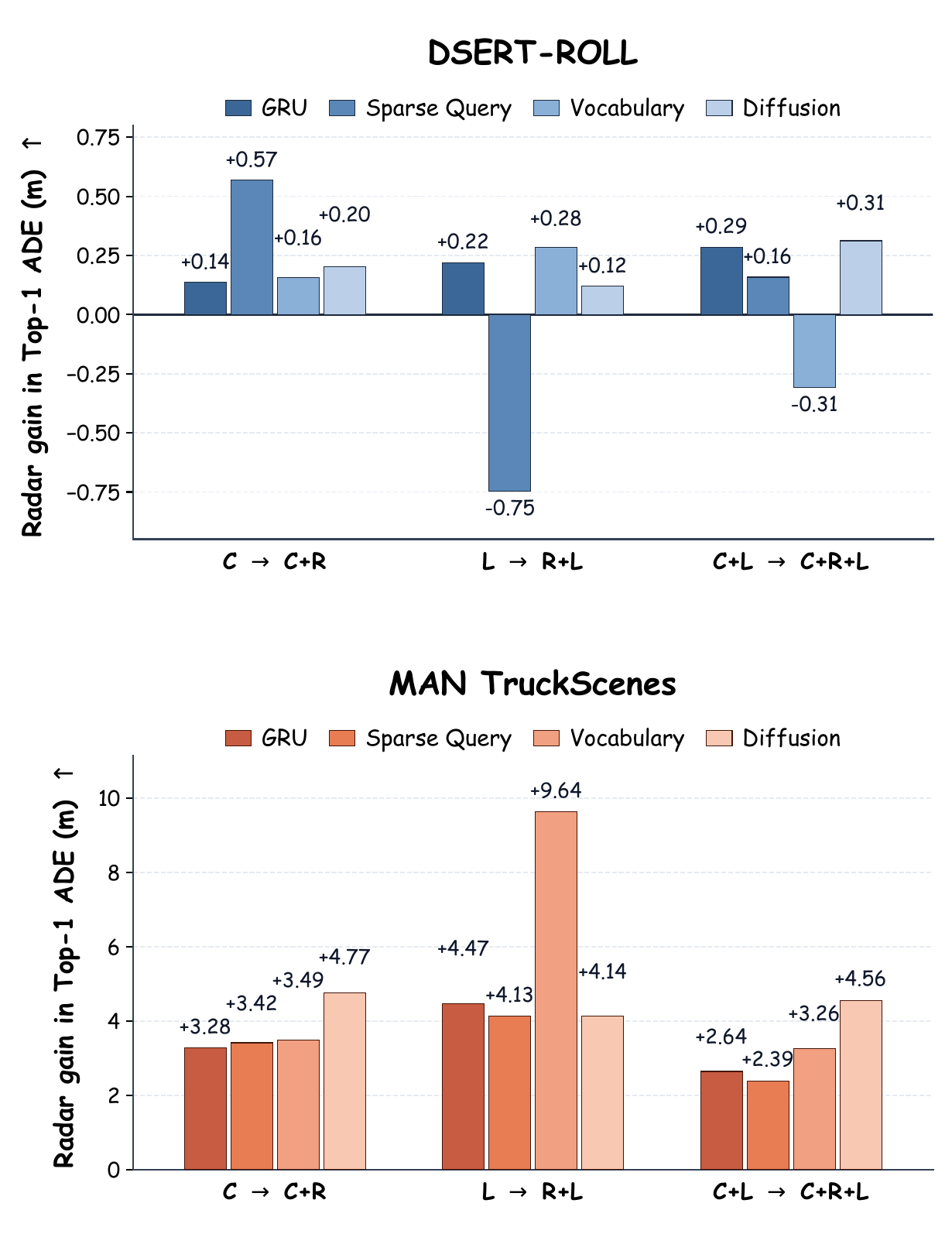}
    \caption{Matched 4D radar gain in six-second Top1ADE on DSERT-RoLL and MAN TruckScenes.
    For each planning baseline, the three comparisons add radar to camera, LiDAR, and camera--LiDAR inputs, respectively.
    Radar gain is defined as $A^{(f)}(X)-A^{(f)}(X{+}R)$ according to Eq.~\eqref{eq:radar_gain}, so positive values indicate lower planning error after adding radar.}
    \label{fig:radar_gain}
\end{figure}

\subsection{Training Protocol and Implementation}
\label{sec:experimental_setup}

We train every configuration for 80 epochs using AdamW.
We set the learning rate to $10^{-4}$, the weight decay to $10^{-4}$, and the batch size to 16.
All runs use seed 42 without a learning-rate scheduler.

Camera inputs are resized to $192\times288$.
Each sample retains at most 2048 radar points and 4096 LiDAR points.
The camera branch uses the current synchronized images.
The radar and LiDAR branches combine the current point cloud with one preceding frame.
The models receive a six-step ego-motion history sampled at 5\,Hz.
Each 11-D history token encodes relative planar displacement, relative yaw, its sine and cosine, speed, and yaw rate.
It also includes longitudinal and lateral acceleration, curvature, and age.
The current ego-state vector contains speed, yaw, global $x/y$ position, yaw rate, and three zero-padded entries.

Each planning baseline retains its baseline-specific training objective.
The TransFuser-style GRU baseline uses an L1 loss on waypoint positions.
The SparseDrive-style baseline combines a winner-mode increment L1 loss with a $0.5$-weighted sigmoid focal loss for mode scoring.
The VADv2-style baseline uses cross-entropy over the nearest fixed trajectory prototype.
The DiffusionDrive-style baseline combines scene-scoring cross-entropy with an L1 trajectory loss weighted by 8.
We clip its trajectory residuals at 5\,m and exclude targets whose six-second endpoint exceeds 200\,m.

Each baseline uses one checkpoint-selection rule across all sensor configurations.
The GRU, sparse-query, and vocabulary baselines select the checkpoint with the lowest baseline-specific validation loss.
The diffusion baseline selects the checkpoint with the lowest internal-development Top1ADE after its mode cross-entropy passes the configured gate.
\subsection{4D Radar as a Planning Modality}
\label{sec:radar_planning_modality}
\subsubsection{Standalone Planning Capability}
Table~\ref{tab:overall_results} compares four planning baselines under seven sensor configurations.
Radar-only input achieves the lowest single-modality Top1ADE for every planning baseline on both datasets.
On DSERT-RoLL, radar reduces the error relative to camera and LiDAR by $0.16$--$0.56$\,m and $0.02$--$0.16$\,m, respectively; on MAN TruckScenes, the corresponding reductions are $3.44$--$4.85$\,m and $4.13$--$9.70$\,m.
These results establish 4D radar as a consistently effective standalone modality across the two driving domains and four trajectory-generation mechanisms.

\subsubsection{Multi-modal Complementarity}
To quantify the contribution of 4D radar when combined with other sensing modalities, we define the radar gain for planning baseline $f$ and base sensor configuration $X$ as
\begin{equation}
G_R^{(f)}(X)
=
A^{(f)}(X)-A^{(f)}(X{+}R),
\qquad
X\in\{C,L,C{+}L\},
\label{eq:radar_gain}
\end{equation}
where $A^{(f)}$ denotes the six-second Top1ADE.
A positive value indicates a lower error after radar is added.

Fig.~\ref{fig:radar_gain} summarizes the effect of adding radar while keeping the dataset, planning baseline, and the remaining sensing modalities unchanged.
Across the two datasets, radar reduces Top1ADE in 22 of the 24 comparisons, corresponding to three base sensor configurations evaluated with four planning baselines on both datasets. On DSERT-RoLL, adding radar to camera input ($C\rightarrow C{+}R$) reduces Top1ADE for all four planning baselines by $0.14$--$0.57$\,m.
Adding radar to LiDAR ($L\rightarrow R{+}L$) and to camera-LiDAR input ($C{+}L\rightarrow C{+}R{+}L$) improves three of the four baselines in each case.
The two exceptions occur for the sparse-query planner with $L\rightarrow R{+}L$ and the trajectory-vocabulary planner with $C{+}L\rightarrow C{+}R{+}L$, where Top1ADE increases by $0.75$\,m and $0.31$\,m, respectively.
These regressions may reflect planner-specific fusion limitations, where additional radar features perturb sparse-query attention or fixed-vocabulary ranking.
In MAN TruckScenes, adding radar reduces Top1ADE for all 12 comparisons.
The reductions range from $3.28$ to $4.77$\,m for $C\rightarrow C{+}R$, from $4.13$ to $9.64$\,m for $L\rightarrow R{+}L$, and from $2.39$ to $4.56$\,m for $C{+}L\rightarrow C{+}R{+}L$.
In general, 4D radar provides the most consistent improvement when complementing camera sensing, while its benefit in LiDAR-based multi-modal configurations can vary with the planning architecture and dataset.

\subsection{Planning Robustness under Adverse Conditions}
\label{sec:condition_robustness}

Fig.~\ref{fig:condition_specific_radar} reports condition-stratified results for six DSERT-RoLL weather groups and six metadata-derived MAN TruckScenes condition groups.
On DSERT-RoLL, radar-only input outperforms camera-only input in 19 of the 20 adverse-weather baseline-condition pairs; adding radar to camera also improves 19 of the 20 pairs.
The largest gains occur under fog and heavy snow: $R$ reduces Top1ADE relative to $C$ by $0.43$--$4.72$\,m under fog and $6.35$--$15.72$\,m under heavy snow, while $C{+}R$ reduces it by $0.40$--$3.89$\,m and $4.18$--$14.53$\,m, respectively.
On MAN TruckScenes, both $R$ and $C{+}R$ outperform $C$ in all 24 planner--condition comparisons across Clear, Overcast, Rain, Illumination, Twilight, and Dark subsets.
The gains are particularly pronounced under adverse weather and challenging illumination, providing direct evidence that the perceptual robustness of 4D radar can translate into improved downstream planning.
Fig.~\ref{fig:qualitative_planning} visualizes representative planning predictions and their corresponding BEV trajectories using Sparse Query on DSERT-RoLL and GRU on MAN TruckScenes.

\subsection{Discussion}
\label{sec:experiment_discussion}

The experimental results reveal two complementary roles of 4D radar in ego-trajectory planning.
As the only sensing modality, the 4D radar enables strong and consistent planning performance across different driving domains and trajectory-generation mechanisms.
When combined with other sensing modalities, its contribution depends on how effectively the planning architecture integrates heterogeneous sensor features.
The particularly large improvements under adverse weather and challenging illumination indicate that 4D radar is especially valuable when camera or LiDAR observations become less reliable.
At the same time, the two DSERT-RoLL cases in which adding radar increases Top1ADE show that introducing radar features alone does not guarantee improved planning performance.
These findings further highlight the importance of evaluating 4D radar across multiple sensor configurations and planning architectures.

\section{Conclusion and Limitations}
\label{sec:conclusion}

We introduced \method, a modular benchmark for evaluating real-world 4D radar in open-loop ego-trajectory planning.
Across DSERT-RoLL and MAN TruckScenes, 4D radar provides strong planning performance as a sole sensing modality and generally improves multi-modal planning, with the largest benefits under adverse weather and illumination.
These results show that the sensing robustness of 4D radar can translate into improved downstream planning and motivate controlled evaluation across sensor configurations and planning architectures.
The current evaluation is limited to two road-vehicle datasets and open-loop planning on recorded sequences.
Future work will extend \method to additional datasets, simulation-based closed-loop evaluation, and a broader range of radar-equipped autonomous systems.

\bibliographystyle{IEEEtran}
\bibliography{abbrev,main}

\end{document}